# Band Selection and Classification of Hyperspectral Images by Minimizing Normalized Mutual Information

ELkebir Sarhrouni*
LRIT, FSR, UMV-A
Rabat, Morocco
Email: sarhrouni436@yahoo.fr

Ahmed Hammouch
LRGE, ENSET, UMV-SOUISSI
Rabat, Morocco
Email: hammouch_a@yahoo.com

Driss Aboutajdine
LRIT, FSR, UMV-A
Rabat, Morocco
Email: aboutaj@fsr.ac.ma

*Abstract*—Hyperspectral images (HSI) classification is a high technical remote sensing tool. The main goal is to classify the point of a region. The HIS contains more than a hundred bidirectional measures, called bands (or simply images), of the same region called Ground Truth Map (GT). Unfortunately, some bands contain redundant information, others are affected by the noise, and the high dimensionalities of features make the accuracy of classification lower. All these bands can be important for some applications, but for the classification a small subset of these is relevant. In this paper we use mutual information (MI) to select the relevant bands; and the Normalized Mutual Information coefficient to avoid and control redundant ones. This is a feature selection scheme and a Filter strategy. We establish this study on HSI AVIRIS 92AV3C. This is effectiveness, and fast scheme to control redundancy.

## I. INTRODUCTION

In the feature classification domain, the choice of data affects widely the results. For the Hyperspectral image, the bands dont all contain the information; some bands are irrelevant like those affected by various atmospheric effects, see Figure.4, and decrease the classification accuracy. Also there exist redundant bands to complicate the learning system and product incorrect prediction [14]. Even the bands contain enough information about the scene they may can't predict the classes correctly if the dimension of space images, see Figure.3, is so large that needs many cases to detect the relationship between the bands and the scene (Hughes phenomenon) [10]. We can reduce the dimensionality of hyperspectral images by selecting only the relevant bands (feature selection or subset selection methodology), or extracting, from the original bands, new bands containing the maximal information about the classes, using any functions, logical or numerical (feature extraction methodology) [11][9]. Here we focus on the feature selection using mutual information. Hyperspectral images have three advantages regarding the multispectral images [6], see Figure.1:

First: the hyperspectral image contains more than a hundred images but the multispectral contains three at ten images.
Second: hyperspectral image has a spectral resolution (the central wavelength divided by de width of spectral band) about a hundred, but that of multispectral is about ten.
Third: the bands of a hyperspectral image is regularly spaced, those of multispectral image is large and irregularly spaced.
Comment: when we reduce hyperspectral images dimensionality, we must save the precision and high discrimination of substances given by hyperspectral image.

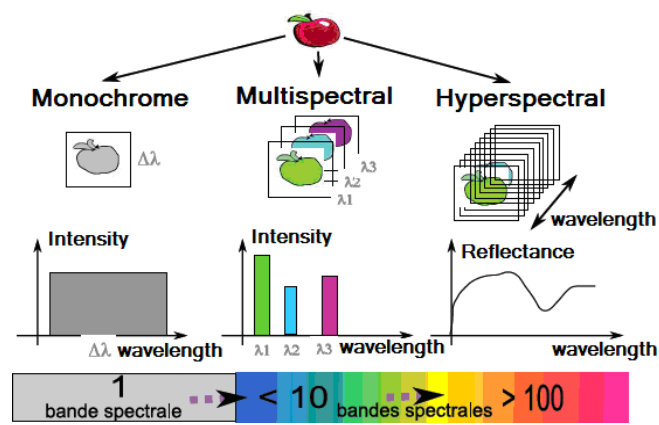

Fig. 1. Precision an dicrimination added by hyperspectral images

In this paper we use the Hyperspectral image AVIRIS 92AV3C (Airborne Visible Infrared Imaging Spectrometer). [2]. It contains 220 images taken on the region "Indiana Pine" at "north-western Indiana", USA [1]. The 220 called bands are taken between 0.4 µm and 2.5 µm. Each band has 145 lines and 145 columns. The ground truth map is also provided, but only 10366 pixels are labeled from 1 to 16. Each label indicates one from 16 classes. The zeros indicate pixels how are not classified yet, Figure.2. The hyperspectral image AVIRIS 92AV3C contains numbers between 955 and 9406. Each pixel of the ground truth map has a set of 220 numbers (measures) along the hyperspectral image. This numbers (measures) represent the reflectance of the pixel in each band. So, the pixel is shown as vector off 220 components, see Figure .3.

Figure.3 shows the vector pixels notion [7]. So, reducing dimensionality means selecting only the dimensions caring a lot of information regarding the classes.

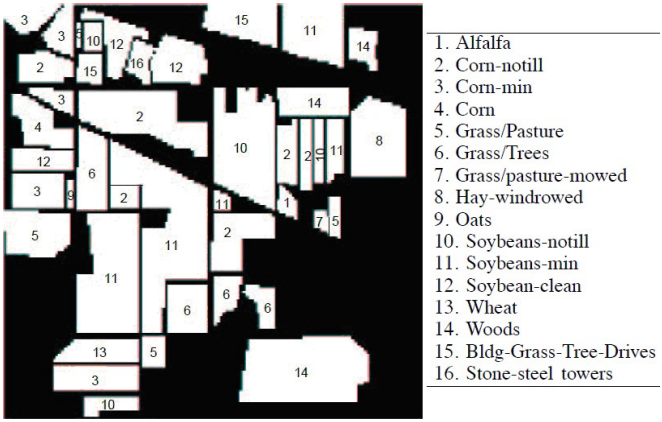

Fig. 2. The Ground Truth map of AVIRIS 92AV3C and the 16 classes

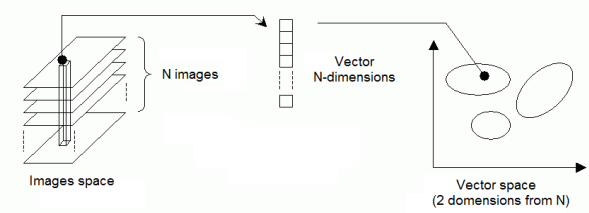

Fig. 3. The notion of pixel vector

We can also note that not all classes are carrier of information. In Figure.4, for example, we can show the effects of atmospheric affects on bands: 155, 220 and other bands. This hyperspectral image presents the problematic of dimensionality reduction.

## II. MUTUAL INFORMAYION BASED SELECTION

In this paragraph we inspect a recent method called band selection scheme using mutual information, and a rejection bandwidth algorithm to eliminate redundancy [3]. [7].

### A. Definition of Mutual Information

This is a measure of exchanged information between tow ensembles of random variables A and B:

$$I(A,B) = \sum log_2 \, p(A,B) \frac{p(A,B)}{p(A).p(B)}$$

Considering the ground truth map, and bands as ensembles of random variables, we calculate their interdependence. Fano [14] has demonstrated that as soon as mutual information of already selected feature has high value, the error probability of classification is decreasing, according to the formula bellow:

$$\frac{H(C/X) - 1}{Log_2(N_c)} \leq P_e \leq \frac{H(C/X)}{Log_2}$$

with :

$$\frac{H(C/X) - 1}{Log_2(N_c)} = \frac{H(C) - I(C;X) - 1}{Log_2(N_c)}$$

and :

$$P_e \leq \frac{H(C) - I(C;X)}{Log_2} = \frac{H(C/X)}{Log_2}$$

The expression of conditional entropy $H(C/X)$ is calculated between the ground truth map (i.e. the classes $C$) and the subset of bands candidates $X$. $N_c$ is the number of classes. So when the features $X$ have a higher value of mutual information with the ground truth map, (is more near to the ground truth map), the error probability will be lower. But it's difficult to compute this conjoint mutual information $I(C;X)$, regarding the high dimensionality [14].

The figure .5 shows the MI between the GT and the real bands of HIS AVIRIS 92AV3C [1].

Many studies use a threshold to choose the relevant bands. Guo [3] use the mutual information to select the top-ranking band, and a filter based algorithm to decide if their neighbours are redundant or not. Sarhrouni et al. [17] use also a filter strategy-based algorithm on MI to select bands. An wrapper strategy based algorithm on MI, Sarhrouni et al. [18], is also introduced.

### B. Bands Selection Scheme

From the mutual information curve, we can make threshold, and we retain the bands that have mutual information value above threshold. But the adjacent bands are possibly redundant. Geo [3] propose an algorithm to eliminate redundancy. It's described in [3] as following: "Let $S$ be the band that maximizes the mutual information. $B_m$ the bandwith, $N$ the number of $S$ neighboring bands.

First define $d(n) = \Delta(MI(n) - MI(n-1))$. Let $X$ be the number of bands to be selected. At some point in the selection process, let $SS$ be the set of selected bands, and let $R$ be the set of remaining bands. We initialize the process with SS={} , and $R = \{1...220\}$".

---

Algorithm 1 : Extracted from[3].

$S = argmax_s(IM\ (s))$
$N = \{S - B_m...S..S + B_m\}$

$d(n) = \{\Delta(MI(n) - MI(n-1), n \in N)\}$.
while $|SS| < X$ do
  if $max_n d(n) < Threshold$ then
    $SS \leftarrow SS \cup S$ ; and $R \leftarrow R \setminus S \setminus N$ ;
  else
    $SS \leftarrow SS \cup S$ ; and $R \leftarrow R \setminus S$ ;
  end if
end while

---

For more detail see to [3].

## C. Discussion and Critics of this Method

This algorithm is applied on mutual information calculated with the estimated ground truth map Figure.4. Like at [3] 50% of the labeled pixels are randomly chosen and used in training; and the other 50% are used for testing classification. The classifier used is the SVM [5] [12] [4]. The classification accuracy is more than 90% for more than 50 bands retained. But the number of neighbors and the threshold must manually adjust with no reason. The most inconvenient of this method is how it measures redundancy: small values of $d(n) = (MI(n)-MI(n-1))$ doesn't necessary an expression of redundancy. It's seed at [3] that the neighboring bands are possibly redundant. So with this method, the advantage (the precision viewed at section I) of hyperspectral images regarding the multispectral images is avoid, because the precious information can be avoided.

## D. Partial conclusion

In this section we inspect the effectiveness of mutual information based selection for hyperspectral images. In the next step we use also the mutual information.

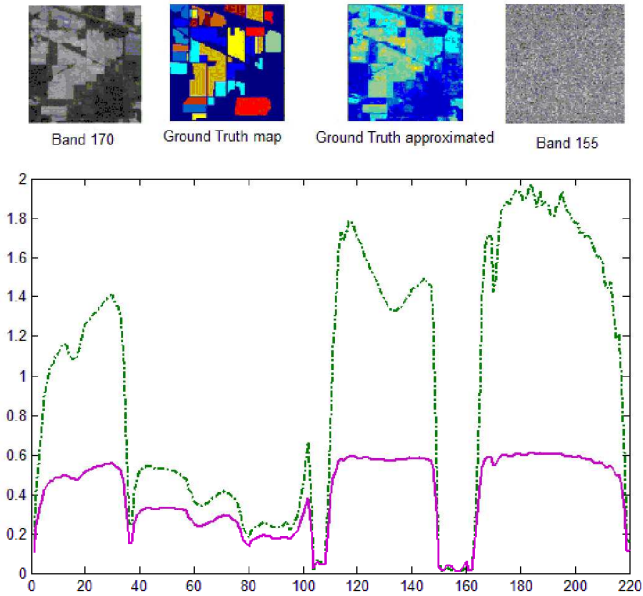

Fig. 4. Mutual information of AVIRIS with the Ground Truth map (solid line) and with the ground approximated by averaging bands 170 to 210 (dashed line) .

## III. PRINCIPE OF THE PROPOSED METHOD

For this section we synthesize 19 bands from the GT, see Figure.2, by adding noise, cutting some substances etc. see Figure.5. Each band has 145 lines and 145 columns. The ground truth map is also provided, but only 10366 pixels are labelled from 1 to 16. Each label indicates one from 16 classes. The zeros indicate pixels how are not classified yet, Figure.2. We can show the Mutual information of GT and the synthetic bands at Figure.6.

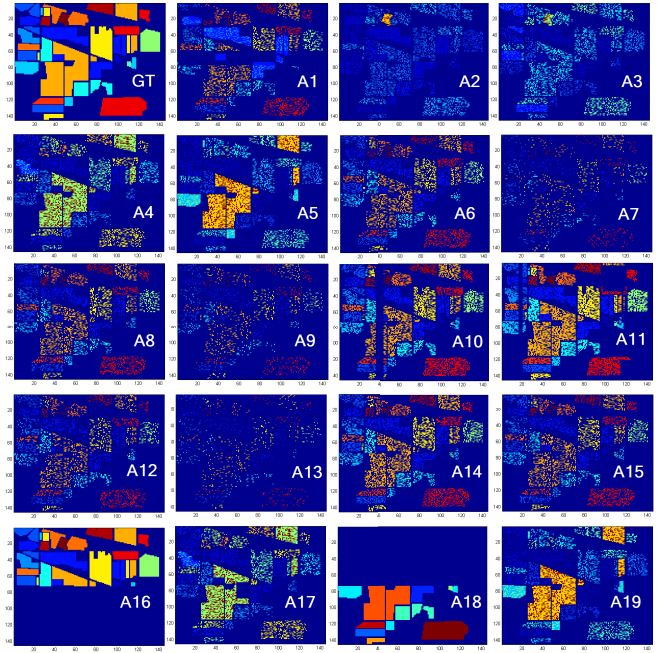

Fig. 5. The synthetic bands used for the study.

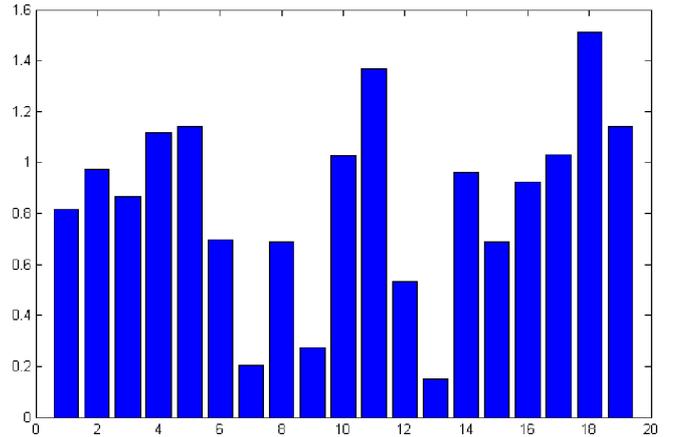

Fig. 6. Mutual information of GT and synthetic bands.

## A. Definition of Normalized Mutual Information

This is one of *normalized* form of Mutual Information; introduced by Witten and Frank 2005 [19]. Its defined as bellow:

$$U(A,B) = \frac{MI(A,B)}{\sqrt[2]{(H(A).H(B))}}$$

*H(X)* is the Entropy of set random variable *X*. Some studies use the *Normalized MI* for recalling images in medical images treatment [9]. Numerous studies use Normalized mutual information [20][21][22].

## B. Principe of Relevant Bands Selection

With a threshold 0.4 of MI calculated in Figure.4, we obtain 16 relevant bands: $A_i$

with $i \in \{1, 2, 3, 4, 5, 6, 8, 10, 11, 12, 14, 15, 16, 17, 18, 19\}$
We can visually verify the resemblance of GT and the bands more informative, bout in synthetic and the real data bands of AVIRIS 92AV3C. See Figure.4 and Figure.5.

## C. Principle to Select no Redundant Bands

First: We order the remaining bands, in increasing order of their MI with the GT. So, we have:
$\{A_{12} A_8 A_{15} A_6 A_1 A_3 A_{14} A_{16} A_2 A_{10} A_{17} A_4 A_{19} A_5 A_{11} A_{18}\}$

Second: We fix a threshold to control redundancy, here 0.7. Then we compute the *Normalized MI*: $U(A_i, A_j)$ for all couple $(i, j)$ of the ensemble:
$S = \{8, 15, 6, 1, 3, 14, 16, 2, 10, 17, 4, 19, 5, 11, 18\}$.

TABLE I
THE NORMALIZED MI OF THE RELEVANT SYNTHETIC BANDS.

| | 12 | 8 | 15 | 6 | 1 | 3 | 16 | 14 | 2 | 10 | 17 | 4 | 19 | 5 | 11 | 18 |
|---|---|---|---|---|---|---|---|---|---|---|---|---|---|---|---|---|
| 12 | 1 | 0,13 | 0,14 | 0,12 | 0,16 | 0,15 | 0,17 | 0,16 | 0,17 | 0,18 | 0,16 | 0,17 | 0,18 | 0,18 | 0,21 | 0,23 |
| 8 | 0,13 | 1 | 0,16 | 0,16 | 0,18 | 0,18 | 0,20 | 0,20 | 0,20 | 0,20 | 0,20 | 0,21 | 0,20 | 0,20 | 0,24 | 0,27 |
| 15 | 0,14 | 0,16 | 1 | 0,16 | 0,18 | 0,18 | 0,20 | 0,19 | 0,20 | 0,20 | 0,20 | 0,21 | 0,21 | 0,20 | 0,24 | 0,27 |
| 6 | 0,12 | 0,16 | 0,16 | 1 | 0,18 | 0,19 | 0,21 | 0,19 | 0,20 | 0,20 | 0,19 | 0,21 | 0,22 | 0,21 | 0,24 | 0,27 |
| 1 | 0,16 | 0,18 | 0,18 | 0,18 | 1 | 0,19 | 0,22 | 0,22 | 0,35 | 0,23 | 0,21 | 0,22 | 0,21 | 0,20 | 0,29 | 0,33 |
| 3 | 0,15 | 0,19 | 0,18 | 0,19 | 0,19 | 1 | 0,21 | 0,23 | 0,24 | 0,21 | 0,22 | 0,23 | 0,22 | 0,28 | 0,34 |
| 16 | 0,17 | 0,20 | 0,20 | 0,20 | 0,22 | 0,21 | 1 | 0,25 | 0,22 | 0,25 | 0,26 | 0,30 | 0,35 | 0,35 | 0,33 | 0,07 |
| 14 | 0,16 | 0,20 | 0,19 | 0,19 | 0,22 | 0,23 | 0,25 | 1 | 0,24 | 0,25 | 0,25 | 0,27 | 0,28 | 0,26 | 0,30 | 0,33 |
| 2 | 0,17 | 0,20 | 0,20 | 0,20 | 0,35 | 0,23 | 0,22 | 0,24 | 1 | 0,25 | 0,23 | 0,24 | 0,25 | 0,24 | 0,32 | 0,37 |
| 10 | 0,18 | 0,20 | 0,20 | 0,20 | 0,23 | 0,24 | 0,25 | 0,25 | 0,25 | 1 | 0,25 | 0,27 | 0,26 | 0,25 | 0,30 | 0,36 |
| 17 | 0,16 | 0,21 | 0,20 | 0,19 | 0,21 | 0,21 | 0,26 | 0,25 | 0,23 | 0,25 | 1 | 0,95 | 0,38 | 0,35 | 0,30 | 0,34 |
| 4 | 0,17 | 0,22 | 0,21 | 0,21 | 0,22 | 0,22 | 0,30 | 0,27 | 0,24 | 0,27 | 0,95 | 1 | 0,43 | 0,41 | 0,32 | 0,34 |
| 19 | 0,18 | 0,21 | 0,21 | 0,22 | 0,21 | 0,23 | 0,35 | 0,28 | 0,25 | 0,26 | 0,38 | 0,43 | 1 | 0,97 | 0,33 | 0,30 |
| 5 | 0,18 | 0,20 | 0,20 | 0,21 | 0,20 | 0,22 | 0,36 | 0,26 | 0,24 | 0,25 | 0,35 | 0,40 | 0,97 | 1 | 0,33 | 0,27 |
| 11 | 0,21 | 0,24 | 0,24 | 0,24 | 0,29 | 0,28 | 0,33 | 0,30 | 0,32 | 0,30 | 0,30 | 0,32 | 0,33 | 0,33 | 1 | 0,41 |
| 18 | 0,23 | 0,27 | 0,27 | 0,27 | 0,33 | 0,34 | 0,07 | 0,33 | 0,37 | 0,36 | 0,34 | 0,34 | 0,30 | 0,27 | 0,41 | 1 |

Observation 1: Figure.5 shows that the band $A_{17}$ is practically the same at $A_4$. Table I shows $U(A_{17}, A_4)$ near to 100% (0.95). This indicates a high redundancy.

Observation 2: Figure.5 shows that the bands $A_{16}$ and $A_{18}$ are practically disjoint, i.e., they are not redundant. Table.I. shows $U(A_{16}, A_{18}) = 0.07$. This indicates no redundancy. The ensemble of selected bands became $SS = \{16, 18\}$, and the bands $A_{16}, A_{18}$ will be discarded from the Table.I. Now we can emit this rule:
*Rule: Each band candidate will be added at SS if and only if their Normalized MI values with all elements off SS, are less than the thresholds (here 0.7).*

## D. The Algorithm Used

Algorithm.2 describes the proposed method to treat relevence and redundancy with Normalized MI.

## IV. APPLICATION ON HIS AVIRIS 92AV3C

We apply the proposed method on the hyperspectral image AVIRIS 92AV3C [1], 50% of the labelled pixels are randomly chosen and used in training; and the other 50% are used for testing classification [3][17][18]. The classifier used is the SVM [5][12] [4].

---

**Algorithm 2**: *Band* is the HSI. Let $Th_{relevance}$ the threshold for selecting bands more informative, $Th_{redundancy}$ the threshold for redundancy control.

1) Compute the Mutual Information (*MI*) of the bands and the Ground Truth map.
2) Make bands in ascending order by there *MI* value
3) Cut the bands that have a lower value than the threshold $Th_{relevance}$, the subset remaining is *S*.
4) Initialization: $n = length(S), i = 1$, *D* is a bidirectional array values=1;
//any value greater than 1 can be used, it's useful in step 6)
5) Computation of bidirectional Data $D(n, n)$:
for 1:=1 to n step 1 do
  for j:=1 to n step 1 do
    $D(i, j) = U(Band_{S(i)}, Band_{S(j)})$;
    $D(j, i) = U(Band_{S(j)}, Band_{S(j)})$;
  end for
end for
//Initialization of the Output of the algorithm
6) $SS = \{\}$;
while $min(D) < Th_{redundancy}$ do
  // Pick up the argument of the minimum of D
  $(x, y) = argmin(D(.,.))$;
  if $\forall l \in SS \ D(x, l) < Th_{redundancy}$ then
    // x is not redundant with the already selected bands
    $SS = SS \cup \{x\}$
  end if
  if $\forall l \in SS \ D(l, y) < Th_{redundancy}$ then
    // y is not redundant with the already selected bands
    $SS = SS \cup \{y\}$
  end if
  $D(x, y) = 1$; // The cells $D(x, y)$ will not be checked as minimum again
end while
7) Finish: The final subset SS contains bands according to the couple of thresholds ($Th_{relevance}, T_{redundancy}$).

---

## A. Mutual Information Curve of Bands

We can visually detect that the lowest MI values correspond to the no informative bands, regarding the GT. See Figure.4. Here we apply a thresholding for conserving only most informative bands.

## B. Usage of Normalize MI

From the remaining subset bands, we must eliminate redundant ones using the proposed algorithm. Table II gives the accuracy off classification for a number of bands with several thresholds.

TABLE II
CLASSIFICATION ACCURACY FOR SEVERAL COUPLES OF THRESHOLDS (TH,IM) AND THEIR CORRESPONDING NUMBER OF BANDS RETAINED

| | | TU: Threshold for control the relevance (Normalized MI of bands with GT) | | | | | | | | | | | | | | | |
|---|---|---|---|---|---|---|---|---|---|---|---|---|---|---|---|---|---|
| | | MI = 0 | | MI > 0,4 | | MI > 0,45 | | MI > 0,57 | | MI > 0,6 | | MI > 0,9 | | MI > 0,91 | | MI > 0,94 | |
| | | N.B | ac(%) | N.B | ac(%) | N.B | ac(%) | N.B | ac(%) | N.B | ac(%) | N.B | ac(%) | N.B | ac(%) | N.B | ac(%) |
| TH: Threshold for control the redundancy | 0,10 | 26 | 37,68 | 3 | 46,31 | 3 | 46,62 | - | - | - | - | - | - | - | - | - | - |
| | 0,15 | 30 | 44,40 | 6 | 59,42 | 6 | 60,65 | 2 | 43,74 | 2 | 43,25 | - | - | - | - | - | - |
| | 0,20 | 34 | 45,31 | 10 | 64,70 | 10 | 65,44 | 6 | 59,69 | 5 | 51,61 | - | - | - | - | - | - |
| | 0,25 | 40 | 46,50 | 13 | 68,52 | 13 | 68,71 | 8 | 56,63 | 8 | 61,82 | - | - | - | - | - | - |
| | 0,30 | 47 | 47,50 | 20 | 75,32 | 18 | 74,67 | 15 | 64,62 | 14 | 67,60 | - | - | - | - | - | - |
| | 0,35 | 59 | 47,13 | 29 | 77,77 | 29 | 77,67 | 25 | 73,33 | 23 | 71,03 | 3 | 65,58 | 2 | 55,48 | - | - |
| | 0,40 | 70 | 46,76 | 40 | 81,41 | 38 | 80,34 | 32 | 77,48 | 28 | 75,03 | 5 | 71,73 | 4 | 63,80 | - | - |
| | 0,43 | 78 | 46,35 | 47 | 83,21 | 46 | 82,82 | 37 | 77,17 | 32 | 77,65 | 6 | 73,29 | 4 | 63,80 | - | - |
| | 0,45 | 90 | 45,88 | 56 | 84,08 | 54 | 83,44 | 45 | 80,60 | 41 | 80,40 | 11 | 75,86 | 7 | 65,69 | 2 | 52,09 |
| | 0,46 | 93 | 45,68 | 61 | 84,69 | 58 | 84,32 | 49 | 81,88 | 44 | 80,89 | 13 | 78,28 | 10 | 69,02 | 2 | 52,09 |
| | 0,47 | 102 | 45,14 | 66 | 85,17 | 61 | 84,24 | 52 | 82,10 | 50 | 81,38 | 16 | 79,82 | 12 | 71,75 | 2 | 52,09 |
| | 0,48 | 109 | 45,10 | 70 | 86,17 | 65 | 85,29 | 56 | 82,04 | 54 | 82,14 | 19 | 80,38 | 14 | 72,73 | 3 | 55,19 |
| | 0,49 | 115 | 44,77 | 76 | 86,56 | 71 | 86,17 | 61 | 83,07 | 59 | 82,74 | 23 | 81,39 | 18 | 74,38 | 3 | 55,19 |
| | 0,50 | 121 | 44,54 | 80 | 86,73 | 75 | 86,46 | 64 | 83,93 | 62 | 83,77 | 25 | 81,73 | 20 | 75,04 | 4 | 56,65 |
| | 0,51 | 127 | 44,15 | 86 | 87,46 | 81 | 87,12 | 68 | 84,34 | 65 | 84,04 | 29 | 82,70 | 23 | 75,67 | 6 | 57,82 |
| | 0,52 | 136 | 43,62 | 90 | 87,75 | 85 | 87,49 | 73 | 85,02 | 71 | 85,04 | 30 | 83,28 | 25 | 76,89 | 7 | 58,08 |
| | 0,53 | 141 | 43,46 | 96 | 87,63 | 91 | 87,43 | 75 | 85,21 | 74 | 84,86 | 32 | 82,93 | 27 | 77,36 | 10 | 59,75 |
| | 0,54 | 147 | 43,23 | 104 | 87,94 | 99 | 87,61 | 81 | 85,82 | 79 | 85,41 | 39 | 83,99 | 34 | 78,32 | 13 | 60,78 |
| | 0,55 | 154 | 42,55 | 108 | 87,94 | 103 | 87,78 | 84 | 86,23 | 82 | 85,68 | 40 | 83,64 | 35 | 78,67 | 15 | 61,84 |
| | 0,56 | 158 | 42,35 | 110 | 87,78 | 105 | 87,63 | 85 | 86,07 | 83 | 86,11 | 42 | 84,16 | 36 | 78,49 | 16 | 61,89 |
| | 0,70 | 220 | 38,96 | 173 | 88,72 | 163 | 88,41 | 128 | 87,88 | 126 | 87,55 | 67 | 86,71 | 54 | 81,77 | 22 | 63,72 |
| | 0,90 | 220 | 38,96 | 173 | 88,72 | 163 | 88,41 | 128 | 87,88 | 126 | 87,55 | 67 | 86,71 | 54 | 81,77 | 22 | 63,72 |
| | 1,00 | 220 | 38,96 | 173 | 88,72 | 163 | 88,41 | 128 | 87,88 | 126 | 87,55 | 67 | 86,71 | 54 | 81,77 | 22 | 63,72 |

N,B : Number of Banbds retained for the couple of threshold (MI,TH)

ac(%) : The accuracy of classification calculated for the couple of threshold (MI,TH)

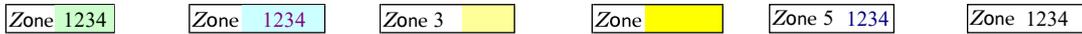

Zone 1234 | Zone 1234 | Zone 3 | Zone | Zone 5 1234 | Zone 1234

## C. Results

Results in Table.II allow us to distinguish six zones of couple values of thresholds (TH,IM):
Zone1: This is practically no control of relevance and no control of redundancy. So there is no action of the algorithm.
Zone2: This is a hard selection: a few more relevant and no redundant bands are selected.
Zone3: This is an interesting zone. We can have easily 80% of classification accuracy with about 40 bands.
Zone4: This is the very important zone; we have the very useful behaviors of the algorithm. For example, with a few numbers of 19 bands we have the classification accuracy 80%.
Zone5: Here we make a hard control of redundancy, but the bands candidates are nearer to the GT, and they my be more redundant. So we can't have interesting results.
Zone6: When we do not control properly the relevance, some bands affected bay noise may be non- redundant, and can be selected, so the accuracy of classification is decreasing.

Partial conclusion: This algorithm is very effectiveness for redundancy and relevance control, in feature selection area.

The most difference of this algorithm regarding previous works is the separation of the tow process: avoiding redundancy and selecting more informative bands. Sarhrouni et al. [17] use also a filter strategy based algorithm on MI to select bands and an another wrapper strategy algorithm also based on MI [18], Guo[3] used a filter strategy with threshold control redundancy, but in those works, the tow process, i.e avoiding redundancy and avoiding no informative bands, are made at the same time by the same threshold.

## V. CONCLUSION

Until recently, in the data mining field, and features selection in high dimensionality, the problematic is always open. Some heuristic methods and algorithms have to select relevant and no redundant subset features. In this paper we introduce an algorithm in order to process separately the relevance and the redundancy. We apply our method to classify the region "Indiana Pine" with the Hyperspectral Image AVIRIS 92AV3C. This algorithm is a Filter strategy (i.e. with no call to classifier during the selection). In the first step we use mutual information to pick up relevant bands by thresholding (like most of methods already used). The second step introduces a new algorithm to measure redundancy with Symmetric Uncertainty coefficient. We conclude the effectiveness of our method and algorithm the select the relevant and no redundant bands. This algorithm allows us a wide area of possible fasted applications. But the question is always open: no guaranties that the chosen bands are the optimal ones; because some redundancy can be important to reinforcement of learning classification system. So the thresholds controlling relevance redundancy is a very useful tool to caliber the selection, in real time application. This is a very positive point for our algorithm; it can be implemented in a real time application, because in commercial applications, the inexpensive filtering algorithms are urgently preferred.